\documentclass{article}
\usepackage{ijcai22}

\usepackage{times}

\usepackage{soul}
\usepackage{url}
\usepackage[hidelinks]{hyperref}
\usepackage[utf8]{inputenc}
\usepackage[small]{caption}
\usepackage{graphicx}
\usepackage{amsmath}
\usepackage{booktabs}
\usepackage{bm}
\usepackage{amssymb}
\usepackage{booktabs}
\usepackage{multirow}
\usepackage{subfigure}

\title{Relational Triple Extraction: One Step is Enough}

\author{
	Yu-Ming Shang$^1$
	\and
	Heyan Huang$^{1}$\footnote{Corresponding author}\and
	Xin Sun$^{1}$\and
	Wei Wei$^2$\And
	Xian-Ling Mao$^1$
	\affiliations
	$^1$School of Computer Science \& Technology, Beijing Institute of Technology, Beijing, China\\
	$^2$Huazhong University of Science and Technology, Hu’bei, China\\
	\emails
	\{ymshang, hhy63, sunxin\}@bit.edu.cn,
	Weiw@hust.edu.cn,
	maoxl@bit.edu.cn
}

\begin{document}

\maketitle

\begin{abstract}
	
	Extracting relational triples from unstructured text is an essential task in natural language processing and knowledge graph construction.
	Existing approaches usually contain two fundamental steps: (1) finding the boundary positions of head and tail entities; (2) concatenating specific tokens to form triples.
	However, nearly all previous methods suffer from the problem of error accumulation, i.e., the boundary recognition error of each entity in step (1) will be accumulated into the final combined triples.	
	To solve the problem, in this paper, we introduce a fresh perspective to revisit the triple extraction task, and propose a simple but effective model, named DirectRel. 
	Specifically,
	the proposed model first generates candidate entities through enumerating token sequences in a sentence, and then transforms the triple extraction task into a linking problem on a ``head $\rightarrow$ tail" bipartite graph.
	By doing so, all triples can be directly extracted in only one step.
	Extensive experimental results on two widely used datasets demonstrate that the proposed model performs better than the state-of-the-art baselines.

\end{abstract}

\section{Introduction}

	Relational triple extraction, defined as the task of extracting pairs of entities and their relations in the form of (head, relation, tail) or ($h, r, t$) from unstructured text, is an important task in natural language processing and automatic knowledge graph construction.
	Traditional pipeline approaches \cite{zelenko2003,chan-roth-2011-exploiting} separate this task into two independent sub-tasks: entity recognition and relation classification while ignoring their intimate connections. Thus, they suffer from the error propagation problem. To tackle this problem, recent studies focus on exploring joint models to extract relational triples in an end-to-end manner.
	
	According to their differences in the extraction procedure, existing joint methods can be broadly divided into three categories: sequence labeling,  table filling and text generation. 
	Sequence labeling methods \cite{zheng-etal-2017-joint,sun2019,yuan2020,wei-etal-2020-novel,zheng-2021-prgc,ren2021simple} utilize various tagging sequences to determine the start and end position of entities, sometimes also including relations.
	Table filling methods \cite{wang-etal-2020-tplinker,yan-etal-2021-partition} construct a table for a sentence and fill each table cell with the tag of the corresponding token-pair.
	Text generation methods \cite{zeng-etal-2018-extracting,zeng-2020-CopyMTL,sui2020joint,ye-2021-contrasive} treat a triple as a token sequence, and employ encoder-decoder architecture to generate triple elements like machine translation.
	
	Although these methods have achieved promising success, most of them suffer from the same problem: error accumulation.
	Concretely, these methods need to first determine the start and end position of head and tail entities, then splice the corresponding tokens within the entity boundaries to form triples.
	Unfortunately, the identification of each boundary token may produce errors, which will be accumulated into the predicted triples.
	As a result, once the recognition of one boundary token fails, the extraction of all triples associated with this token will fail accordingly.
	
	\begin{figure}[t]
		\centering
		\includegraphics[width=1\columnwidth]{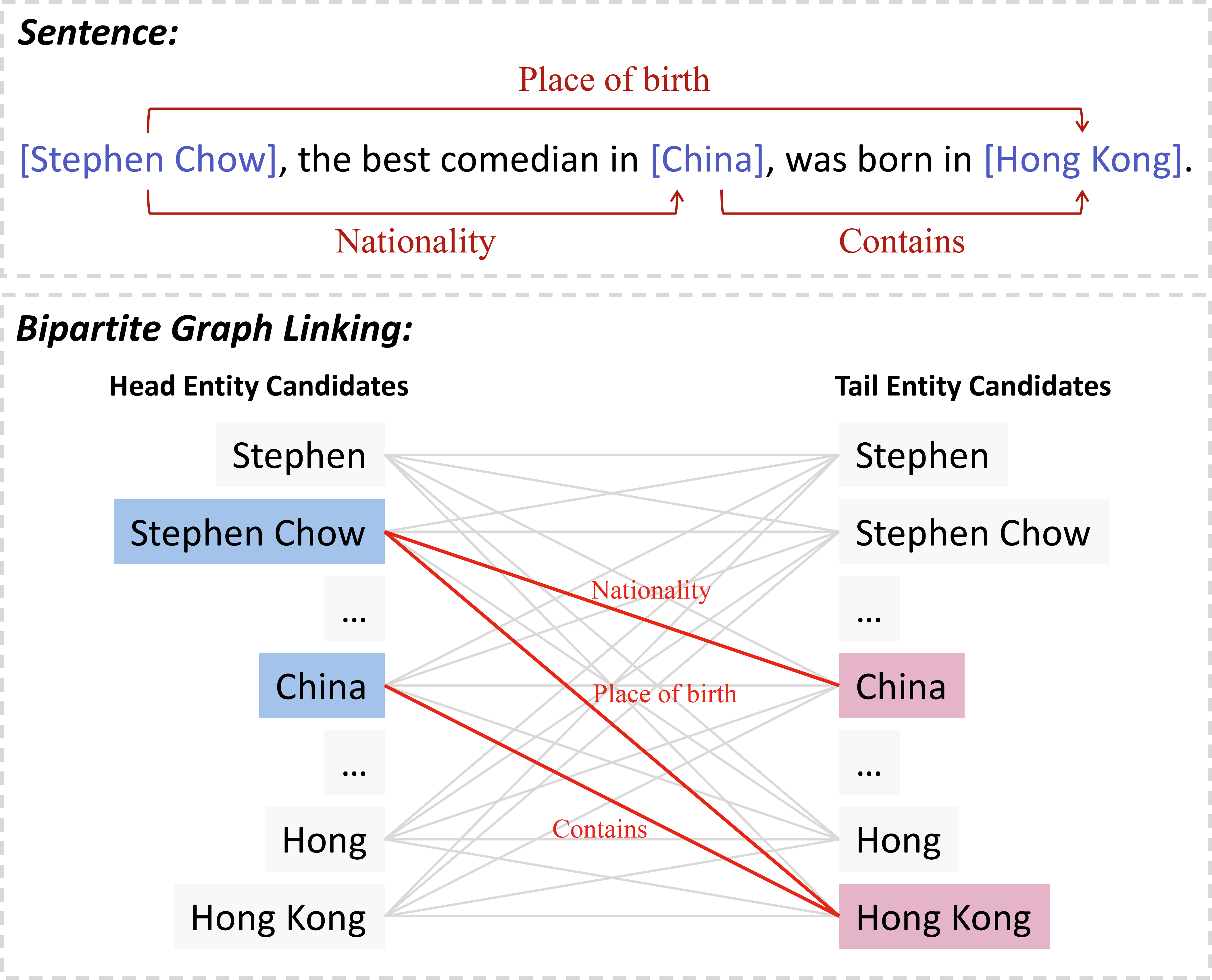} 
		\caption{
		   An example of bipartite graph linking based triple extraction. We enumerate all token sequences of length less than 2 as candidate entities.
		}
		\label{fig:example}
	\end{figure}
	
	Intuitively, if we can directly extract relational triples from unstructured sentences in a one-step operation without identifying the boundary tokens of entities, the above problem will be solved.
	Following this intuition, we revisit the triple extraction task from a new perspective --- bipartite graph linking.
	As shown in Figure \ref{fig:example}, it can be observed that an entity is essentially composed of several consecutive tokens. In other words, if we exhaustively enumerate token sequences of a sentence, the result must contain all correct entities.
	Thus, the triple (\textit{Stephen Chow}, \texttt{Nationality}, \textit{China}) can be directly identified by predicting whether there is a ``link" \texttt{Nationality} between the two candidate entities ``\textit{Stephen Chow}" and ``\textit{China}".

	Inspired by the above intuition, in this paper, we propose a novel relational triple extraction model, named DirectRel, which is able to directly extract all triples from unstructured text in one step.
	Specifically,
	given a sentence, we first generate candidate entities by enumerating token sequences during data pre-processing.
	Then, we design a link matrix for each relation to detect whether two candidate entities can form a valid triple, and transform triple extraction into a relation-specific bipartite graph linking problem.
	Obviously, such a solution would generate redundant negative samples during the training phase. 
	To address this issue, DirectRel conducts downsampling on negative entities during training.
	Extensive experimental results demonstrate that DirectRel outperforms the state-of-the-art approaches on two widely used benchmarks.	
	
	In summary, the main contributions of this paper are as follows:
	
	\begin{itemize}
		
		\item We propose a novel perspective to transform the relational triple extraction task into a bipartite graph linking problem, which addresses the error accumulation issue from design.
			
		\item As far as we know, the proposed DirectRel is the first model that is capable of directly extracting all relational triples from unstructured text with one-step computational logic.
		
		\item We conduct extensive experiments on two widely used datasets, and the results indicate that our model performs better than state-of-the-art baselines.
				
	\end{itemize}
	
\section{Related Work}
	
	This paper focuses on the joint extraction of relational triples from sentences. Related works can be roughly divided into three categories.
	
	The first category is sequence labeling methods, which transform the triple extraction task into several interrelated sequence labeling problems. 
	For example, a classical method NovelTagging \cite{zheng-etal-2017-joint} designs a complex tagging scheme, which contains the information of entity beginning position, entity end position and relation.
	Some studies \cite{sun2019,liu2020,yuan2020} first use sequence labeling to identify all entities in a sentence, and then perform relation detection through various classification networks.
	Recently, \citeauthor{wei-etal-2020-novel} \shortcite{wei-etal-2020-novel} present CasRel, which first identifies all possible head entities, then for each head entity, applies relation-specific sequence taggers to identify the corresponding tail entities.
	PRGC \cite{zheng-2021-prgc} designs a component to predict potential relations, which constrains the following entity recognition to the predicted relation subset rather than all relations.
	BiRTE \cite{ren2021simple} proposes a bidirectional entity extraction framework to consider \textit{head-tail} and \textit{tail-head} extraction order simultaneously.
		
	The second category is table filling methods, which formulate the triple extraction task as a table constituted by the Cartesian product of the input sentence to itself.
	For example, GraphRel \cite{fu-etal-2019-graphrel} takes the interaction between entities and relations into account via a relation-weighted Graph Convolutional Network.
	TPLinker \cite{wang-etal-2020-tplinker} converts triple extraction as a token pair linking problem and introduces a relation-specific handshaking tagging scheme to align the boundary tokens of entity pairs.
	PFN \cite{yan-etal-2021-partition} utilizes a partition filter network, which generates task-specific features jointly to model the interactions between entity recognition and relation classificaiton.
	
	The third category is text generation methods, which treat a triple as a token sequence and employs the encoder-decoder framework to generate triple elements like machine translation.
	For example, CopyRE \cite{zeng-etal-2018-extracting} generates the relation followed by its two corresponding entities with a copy mechanism, but this method can only predict the last word of an entity.
	Thus, CopyMTL \cite{zeng-2020-CopyMTL} employs a multi-task learning framework to address the multi-token entity problem.
	CGT \cite{ye-2021-contrasive} proposes a contrastive triple extraction method with a generative transformer to address the long-term dependence and faithfulness issues.
	R-BPtrNet \cite{chen-etal-2021-jointly} designs a binary pointer network to extract explicit triples and implicit triples.

	However, nearly all existing methods suffer from the error accumulation problem due to possible errors in entity boundary identification. 
	Different from previous methods, DirectRel proposed in this paper transforms the triple extraction task into a bipartite graph linking problem without determining the boundary tokens of entities. Therefore, our method is able to directly extract all triples from unstructured sentences with a one-step linking operation and naturally address the problem of error accumulation.	
\section{Method}
	
\begin{figure*}[t]
		\centering
		\includegraphics[width=16cm]{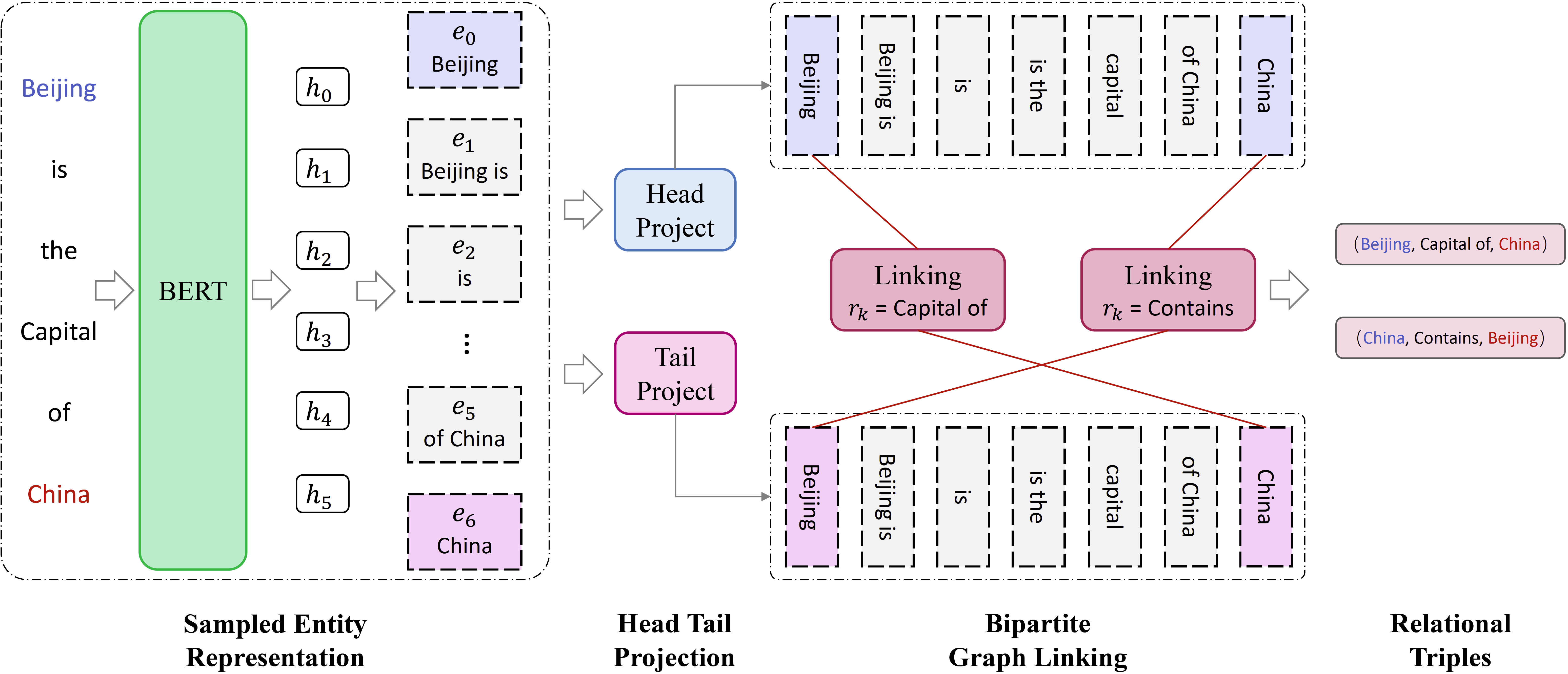} 
		\caption{
		   The architecture of the proposed method, displaying the procedure for handling one sentence that contains two EPO triples (\textit{Beijing}, \texttt{Capital of}, \textit{China}) and (\textit{China}, \texttt{Contains}, \textit{Beijing}). 
		   In this example, the downsampled set $\mathcal{\bar{\mathcal{E}}}$ contains 5 negative entities (marked in grey) and 2 positive entities. 
		   Note that all false links are omitted for convince of illustration. 
		   }
		\label{fig:model}
	\end{figure*}

	The overall architecture of the proposed DirectRel is illustrated in Figure \ref{fig:model}.
	In the following, we first give the task definition and notations in Section \ref{task definition}. 
	Then, the strategies for candidate entities generation are introduced in Section \ref{stragegies}.
	Finally, Section \ref{model details} illustrates the details of the bipartite graph linking based triple extraction.
	
	\subsection{Task Definition}
	\label{task definition}
	
	The goal of relational triple extraction is to identify all possible triples in a given sentence.
	Therefore, the input of our model is a sentence $\mathcal{S} = \{ w_1, w_2, ..., w_L \}$ with $L$ tokens. Its output is a set of triples $ \mathcal{T} = \{ (h, r, t)  | h, t \in \hat{\mathcal{E}}, r_i \in \mathcal{R} \}$, where $\mathcal{R} = \{ r_1, r_2, ..., r_{K} \}$ denotes $K$ pre-defined relations.	
	It is worth noting that, $\hat{\mathcal{E}}$ represents the head and tail entities in triples, not all named entities in the sentence. 
	
	\subsection{Candidate Entities Generation}
	\label{stragegies}
	
	During data pre-processing, we enumerate all consecutive token sequences with length less than $C$ $(C  < L)$ in a sentence as its candidate entities.
	For example, if $C = 2$, the candidate entities of the sentence ``Beijing is the capital of China" are $ \mathcal{E} = \{$ ``\textit{Beijing}", ``\textit{Beijing is}", ``\textit{is}", ``\textit{is the}", ``\textit{the}", ``\textit{the Capital}", ``\textit{Capital}", ``\textit{Capital of}", ``\textit{of}", ``\textit{of China}", ``\textit{China}"$\}$.
	Thus, for a sentence with $L$ tokens, the number of candidate entities $|\mathcal{E}|$ is:
	
	\begin{equation}
		|\mathcal{E}| = L \times C + \frac{C}{2} - \frac{C^2}{2}.
	\end{equation} 
	Obviously, such a strategy will bring two disadvantages:
	First, the training process will bias towards negative triples as they dominate, which will hurt the model's ability to identify positive triples.
	Second, since the number of training sentences is large, too many candidate entities will reduce the training efficiency. 
	To address these issues, for a sentence, we randomly sample $n_{neg}$ negative entities from $ \mathcal{E}$ to train the model together with all ground truth entities, and the new subset is denoted as $\mathcal{\bar{\mathcal{E}}}$.
	
	\subsection{Bipartite Graph Linking}
	\label{model details}
	
	Given a sentence and its candidate entities $\mathcal{\bar{\mathcal{E}}}$, we employ a pre-trained BERT \cite{devlin-etal-2019-bert} as sentence encoder to obtain the $d$-dimensional contextual representation $\bm{h}_i$ for each token:
	
	\begin{equation}
		[ \bm{h}_1, \bm{h}_2, ..., \bm{h}_{L} ] = BERT( [ \bm{x}_1, \bm{x}_2, ..., \bm{x}_{L} ] ),
	\end{equation} 
	where $\bm{x}_i$ is the input representation of the $i$-th token. It is the summation over the corresponding token embedding and positional embedding.
	
	It's worth noting that an entity is usually composed of multiple tokens, to facilitate parallel computation, we need to keep the dimension of different entity representations consistent. Therefore, we take the averaged vector between the start token and end token of entity $e_i \in \bar{\mathcal{E}}$ as its representation:
	
	\begin{equation}
	\label{entity_representation}
		\bm{e}_i = \frac{\bm{h}^{start} + \bm{h}^{end}}{2}.
	\end{equation}
	
	Then, as shown in Figure \ref{fig:model}, we define a directed ``head $\rightarrow$ tail" bipartite graph for triple extraction, which takes the projected entity representations $\bm{E}_{head} = \bm{W}_h^T \bm{E} + \bm{b}_h$ and $\bm{E}_{tail} = \bm{W}_t^T \bm{E} + \bm{b}_t$ as two parts, where $\bm{E}$ is the $d$-dimensional representations of entities obtained by equation (\ref{entity_representation});
	$\bm{W}_h$, $\bm{W}_t$ are two project matrices from token feature space to $d_e$-dimensional head entity space and tail entity space, respectively, allowing the model to identify the head or tail role of each entity; $\bm{b}_{(\cdot)}$ is the bias.
	
	Finally, for each relation $r_k$, we predict the links between every entity pair to determine whether they can form a valid triple:
	
	\begin{equation}
	\label{formula}
		\bm{P}^k = \sigma(\bm{E}_{head}^T \bm{U}_k \bm{E}_{tail}),
	\end{equation}
	where $\sigma$ is the sigmoid activation function, $\bm{U}_k^{d_e \times d_e}$ is a relation-specific link matrix, which models the correlation between two entities with respect to the $k$-th relation.
	The triple $(e_i, r_k, e_j)$ will be treated as correct if the corresponding probability $\bm{P}^k_{ij}$ exceeds a certain threshold $\theta$, or false otherwise.
	Besides, since the entity spans have been determined in the data pre-processing stage, decoding triples from the output of our model becomes easy and straightforward.
	That is, for each relation $r_k$, the predicted triple is ($e_i$\texttt{.span}, $r_k$, $e_j$\texttt{.span}), if $\bm{P}^k_{ij} > \theta$.
	
	\begin{table*}[ht]
		\centering
		\setlength{\tabcolsep}{1.1mm}
		\renewcommand\arraystretch{1.1}
		\begin{tabular}{@{}lccccccccccccccc@{}}
			\toprule[2pt]
			\multicolumn{1}{l}{\multirow{2}{*}{Category}} & \multicolumn{4}{c}{Dataset} & \multicolumn{11}{c}{Details of Test Set}                                          \\ 
			\cmidrule(l){2-5} \cmidrule(l){6-16}
			\multicolumn{1}{l}{}                          & Train    & Valid   & Test   & Relations  & Normal & SEO   & EPO   & HTO & N=1   & N=2 & N=3 & N=4 & N$\geq$5 & Triples  & E-len\\ 
			\midrule
			NYT$^*$                                       & 56,195   & 4,999   & 5,000   & 24   & 3,266  & 1,297 & 978   & 45  & 3,244 & 1,045 & 312   & 291  &108 & 8,110 &   7   \\
			WebNLG$^*$                                    & 5,019    & 500     & 703    & 171    & 245    & 457   & 26    & 84  & 266   & 171   & 131   &90    & 45 & 1,591 & 6  \\
			NYT                                           & 56,195   & 5,000    & 5,000   & 24    & 3,222  & 1,273 & 969   & 117  & 3,240 & 1,047 & 314   & 290  &109 & 8,120 &11  \\
			WebNLG                                        & 5,019    & 500     & 703    & 216   & 239    & 448   & 6     & 85  & 256   & 175   & 138   & 93   & 41 & 1,607  &39 \\ 
			\bottomrule[2pt]
		\end{tabular}
		\caption{Statistics of datasets. $N$ is the number of triples in a sentence, E-len denotes the maximum length of entities using byte pair encoding (BPE), which determines the setting of the hyper-parameter $C$ (the length of candidate entities). }
		\label{tab:dataset}
	\end{table*}

	Obviously, our method can naturally identify nested entities and overlapping triples\footnote{More details about different overlapping patterns are shown in Appendix A.} \cite{zeng-etal-2018-extracting}. 
	Specifically, 
	for nested entity recognition, the correct entities must be included in the candidate entities generated by enumeration.  
	For \textit{EntityPairOverlap} (EPO) case, entity pairs with different relations will be recognized by different relation-specific link matrices.
	For \textit{SingleEntityOverlap} (SEO) case, if two triples have the same relation, there will be two edges in the bipartite graph; if two triples have different relations, they will be identified by different link matrices.
	For \textit{HeadTailOverlap} (HTO) case, the overlapped entity will appear on both parts of the bipartite graph and can also be easily identified.
	
	\subsection{Objective Function}
	
	The objective function of DirectRel is defined as:
	\begin{equation}
		\begin{aligned}
		\mathcal{L} = & - \frac{1}{|\bar{\mathcal{E}}| \times K \times |\bar{\mathcal{E}}|} \times \\
		 & \sum_{i=1}^{|\bar{\mathcal{E}}|}\sum_{k=1}^K\sum_{j=1}^{|\bar{\mathcal{E}}|}(y_t\log(\bm{P}^k_{ij}) + (1-y_t)\log(1 - \bm{P}^k_{ij})),
		\end{aligned}
	\end{equation}
	where $|\bar{\mathcal{E}}|$ is the number of entities used for training, $K$ denotes the number of pre-defined relations, $y_t$ is the gold label of the triple $(e_i, r_k, e_j)$.
		 	 
\section{Experiments}

	Our experiments are designed to evaluate the effectiveness of the proposed DirectRel and analyze its properties. In this section, we first introduce the experimental settings. Then, we present the evaluation results and discussion.
	
	\subsection{Experimental Settings}
	
	\subsubsection{Datasets and Evaluation Metrics}
	
	We conduct experiments on two widely used relational triple extraction benchmarks: NYT \cite{riedel2010} and WebNLG \cite{gardent-etal-2017-creating}.
	
	\begin{itemize}
		
		\item \textbf{NYT}: The dataset is generated by distant supervision, which automatically aligns relational facts in Freebase with the New York Times (NYT) corpus.
		It contains 56k training sentences and 5k test sentences.
		
		\item \textbf{WebNLG}: The dataset is originally developed for Natural Language Generation (NLG) task, which aims to generate corresponding descriptions from given triples. 
		It contains 5k training sentences and 703 test sentences.
	\end{itemize}

	Both NYT and WebNLG have two versions: one version only annotates the last word of entities, denoted as NYT$^*$ and WebNLG$^*$; the other version annotates the whole span of entities, denoted as NYT and WebNLG. Table \ref{tab:dataset} illustrates their detailed statistics.
	Notably, as our model employs byte pair encoding, entities in NYT$^*$ and WebNLG$^*$ may also contain multiple tokens.

	Following previous works \cite{wei-etal-2020-novel,zheng-2021-prgc,ren2021simple}, we adopt standard micro Precision (Prec.), Recall (Rec.) and F1-score (F1) to evaluate the performances. Concretely, a predicted triple ($h, r, t$) is regarded to be correct only if the head $h$, tail $t$ and their relation $r$ are identical to the ground truth. 
	
	\subsubsection{Implementation Details}
	We employ the cased base version\footnote{https://huggingface.co/bert-base-cased} of BERT as sentence encoder. Therefore, the dimension of token representation $\bm{h}_i$ is $d = 768$. The dimension of projected entity representations $d_e$ is set to 900.
	During training, the learning rate is 1e-5, and the batch size is set to 8 on NYT$^*$ and NYT, 6 on WebNLG$^*$ and WebNLG.
	The max length of candidate entities $C$ is 9/6/12/21 on NYT$^*$/WebNLG$^*$/NYT/WebNLG respectively.
	For each sentence, we randomly select $n_{neg} = 100$ negative entities from $\mathcal{E}$ to optimize the objective function of a mini-batch. If there are fewer than 100 candidates in a sentence, all negative entities will be used.
	During inference, we predict links for all candidate entities and the max length $C$ is 7/6/11/20 on NYT$^*$/WebNLG$^*$/NYT/WebNLG respectively\footnote{We present the entity length distribution of datasets in Appendix B, as well as our consideration when setting $C$}. 
	All experiments are conducted on a work station with an AMD 7742 2.25G CPU, 256G Memory, and a single RTX 3090 GPU.
	
	\subsubsection{Baselines}
	We compare our method with the following ten baselines:
	\textbf{GraphRel} \cite{fu-etal-2019-graphrel}, \textbf{MHSA} \cite{yuan2020}, \textbf{RSAN} \cite{liu2020}, \textbf{CopyMTL} \cite{zeng-2020-CopyMTL}, \textbf{CasRel} \cite{wei-etal-2020-novel}, \textbf{TPLinker} \cite{wang-etal-2020-tplinker}, \textbf{CGT} \cite{ye-2021-contrasive}, \textbf{PRGC} \cite{zheng-2021-prgc}, \textbf{R-BPtrNet} \cite{chen-etal-2021-jointly}, \textbf{BiRTE} \cite{ren2021simple}. 
	
	For fair comparison, the reported results for all baselines are directly from the original literature.	
	\begin{table*}[t]
		\centering
		\setlength{\tabcolsep}{2mm}
		\renewcommand\arraystretch{1.1}
		\scalebox{1}{
				\begin{tabular}{@{}lcccccccccccc@{}}
					\toprule[2pt]
					\multicolumn{1}{c}{\multirow{2}{*}{Model}} & \multicolumn{3}{c}{NYT$^*$} & \multicolumn{3}{c}{WebNLG$^*$} & \multicolumn{3}{c}{NYT} & \multicolumn{3}{c}{WebNLG} \\ 
					\cmidrule(l){2-4} \cmidrule(l){5-7} \cmidrule(l){8-10} \cmidrule(l){11-13} 
					\multicolumn{2}{c}{}                       Prec.   & Rec.  & F1    & Prec.    & Rec.   & F1     & Prec.  & Rec.   & F1    & Prec.   & Rec.    & F1   \\  
					\midrule
					GraphRel \cite{fu-etal-2019-graphrel}         & 63.9    & 60.0  & 61.9  & 44.7     & 41.1   & 42.9   & -      & -      & -     & -       & -       & -      \\
					RSAN \cite{yuan2020}            & -       & -     & -     & -        & -      & -      & 85.7  & 83.6  & 84.6 & 80.5   & 83.8   & 82.1  \\ 
					MHSA  \cite{liu2020}  & 88.1   & 78.5 & 83.0 &  89.5 & 86.0 & 87.7 & -   & -   & -  & -   &  -    &  -  \\
					CopyMTL \cite{zeng-2020-CopyMTL}     & -   & -  & -  & -     & -   & -   & 75.7   & 68.7 & 72.0  & 58.0 & 54.9  & 56.4 \\
					\midrule[0.1pt]
					CasRel \cite{wei-etal-2020-novel}          & 89.7    & 89.5  & 89.6  & 93.4     & 90.1   & 91.8   & -      & -      & -     & -       & -       & -      \\
					TPLinker \cite{wang-etal-2020-tplinker}        & 91.3    & 92.5  & 91.9  & 91.8     & 92.0   & 91.9   & 91.4   & 92.6   & 92.0  & 88.9    & 84.5    & 86.7   \\
					CGT  \cite{ye-2021-contrasive}            & \textbf{94.7}    & 84.2  & 89.1  & 92.9     & 75.6   & 83.4   & -      & -      & -     & -       & -       & -      \\
					PRGC  \cite{zheng-2021-prgc}           & 93.3    & 91.9  & 92.6  & 94.0     & 92.1   & 93.0   & 93.5   & 91.9   & 92.7  & 89.9    & 87.2    & 88.5   \\
					R-BPtrNet \cite{chen-etal-2021-jointly}    & 92.7  & 92.5  & 92.6 & 93.7  & 92.8 & 93.3  & - & - & - & - & - & - \\
					BiRTE  \cite{ren2021simple}         & 92.2  & \textbf{93.8}  & 93.0  & 93.2     & 94.0   & 93.6   & 91.9   & \textbf{93.7}   & 92.8   & 89.0    & \textbf{89.5}     & 89.3      \\ 
					\midrule
					DirectRel         & 93.7    & 92.8  & \textbf{93.2}  & \textbf{94.1}     & \textbf{94.1}   & \textbf{94.1}   & \textbf{93.6}& 92.2  & \textbf{92.9} & \textbf{91.0} & 89.0 & \textbf{90.0} \\ 
					\bottomrule[2pt]
				\end{tabular}
		}
		\caption{Precision(\%), Recall (\%) and F1-score (\%) of our proposed DirectRel and baselines. GraphRel, RSAN, MHSA, CopyMTL use LSTM as sentence encoder, while other methods employ a pre-trained BERT to obtain feature representations.}
		\label{tab:main}
	\end{table*}
	
	\begin{table*}[t]
		\setlength\tabcolsep{1mm}
		\renewcommand\arraystretch{1.1}
		\centering
		\scalebox{1}{
			\begin{tabular}{@{}lcccccccccccccccccc@{}}
				\toprule[2pt]
				\multicolumn{1}{c}{\multirow{2}{*}{Model}} & \multicolumn{9}{c}{NYT$^*$}                                                   & \multicolumn{9}{c}{WebNLG$^*$}                                                \\ 
				\cmidrule(l){2-10} \cmidrule(l){11-19} 
				\multicolumn{1}{c}{}                       & Normal & EPO  & SEO  & HTO & N=1  & N=2  & N=3  & N=4  & N$\geq$5 & Normal & EPO  & SEO  & HTO & N=1  & N=2  & N=3  & N=4  & N$\geq$5 \\ 
				\midrule
				CasRel                                     & 87.3   & 92.0 & 91.4 &  77.0$^\S$   & 88.2 & 90.3 & 91.9 & 94.2 & 83.7         & 89.4   & 94.7 & 92.2 &  90.4$^\S$ & 89.3 & 90.8 & 94.2 & 92.4 & 90.9             \\
				TPLinker                                   & 90.1   & 94.0 & 93.4 & 90.1$^\S$  & 90.0 & 92.8 & 93.1 & 96.1 & 90.0             & 87.9   & 95.3 & 92.5 &  86.0$^\S$ & 88.0 & 90.1 & 94.6 & 93.3 & 91.6             \\
				PRGC                                       & 91.0   & 94.5 & 94.0 &  81.8$^\S$  & 91.1 & 93.0 & 93.5 & 95.5 & \textbf{93.0}           & 90.4   & 95.9 & 93.6 &  94.6$^\S$ & 89.9 & 91.6 & 95.0 & 94.8 & 92.8             \\
				BiRTE                                      & 91.4  & 94.2 & \textbf{94.7} & -  & 91.5 & 93.7 & \textbf{93.9} & 95.8 & 92.1                & 90.1     &  94.3   &   \textbf{95.9}  &   - & 90.2 & \textbf{92.9} & 95.7 & 94.6 & 92.0             \\
				\midrule
				DirectRel                                   & \textbf{91.7}   & \textbf{94.8}  & 94.6 &   \textbf{90.0}  & \textbf{91.7} & \textbf{94.1} & 93.5 & \textbf{96.3} & 92.7       &  \textbf{92.0} & \textbf{97.1} & 94.5  & \textbf{94.6} & \textbf{91.6} & 92.2   & \textbf{96.0} & \textbf{95.0} & \textbf{94.9}                \\ 
				\bottomrule[2pt]
			\end{tabular}
		}
		\caption{
		F1-score (\%) on sentences with different overlapping patterns and different triple numbers. 
		$\S$ marks the results reported by PRGC.
		}
		\label{tab:type}
	\end{table*}
	
	\subsection{Results and Analysis}
	\subsubsection{Main Results}
	In Table \ref{tab:main}, we present the comparison results of our DirectRel with ten baselines on two versions of NYT and WebNLG.
	It can be observed that DirectRel outperforms all the ten baselines and achieves the state-of-the-art performance in terms of F1-score on all datasets. 
	Among the ten baselines, CasRel and TPLinker are the representative methods for combining triples through identifying boundary tokens of head and tail entities.
	Our DirectRel outperforms CasRel by 3.6 and 2.3 absolute gains in F1-score on NYT$^*$ and WebNLG$^*$; and outperforms TPLinker by 1.3, 2.2, 0.9, 3.3 absolute gains in term of F1-score on NYT$^*$, WebNLG$^*$, NYT and WebNLG respectively.
	Such results demonstrate that directly extracting entities and relations from unstructured text through a one-step manner can effectively address the problem of error accumulation.
	
	Another meaningful observation is that DirectRel achieves the best F1-score on WebNLG. 
	As mentioned before, the max length of candidate entities on WebNLG is set to 21 during training and 20 during inference. Therefore, each sentence generates a large number of candidate entities, posing a great challenge to our method. 
	Nevertheless, DirectRel achieves the best performance against all baselines, which proves the effectiveness of our strategies of candidate entities generation and negative entities sampling.
	
	\subsubsection{Detailed Results on Complex Scenarios}
	
	To further explore the capability of our DirectRel in handling complex scenarios, we split the test set of NYT$^*$ and WebNLG$^*$ by overlapping patterns and triple number, and the detailed extraction results are shown in Table \ref{tab:type}.
	It can be observed that DirectRel obtains the best F1-score on 13 of the 18 subsets, and the second best F1-score on the remaining 5 subsets. 
	Besides, we can also see that DirectRel obtains more performance gains when extracting EPO triples.
	We attribute the outstanding performance of DirectRel to its two advantages:
	First, it effectively alleviates the error accumulation problem and ensures the precision of extracted triples.	
	Second, it applies a relation-specific linking between every entity pair, guaranteeing the recall of triple extraction.
	Overall, the above results adequately prove that our proposed method is more effective and robust than baselines when dealing with complicated scenarios.
	
	\begin{figure*}
		\centering
		\subfigure[Training Time]{
		\label{speed}
		\includegraphics[width = 5.05cm]{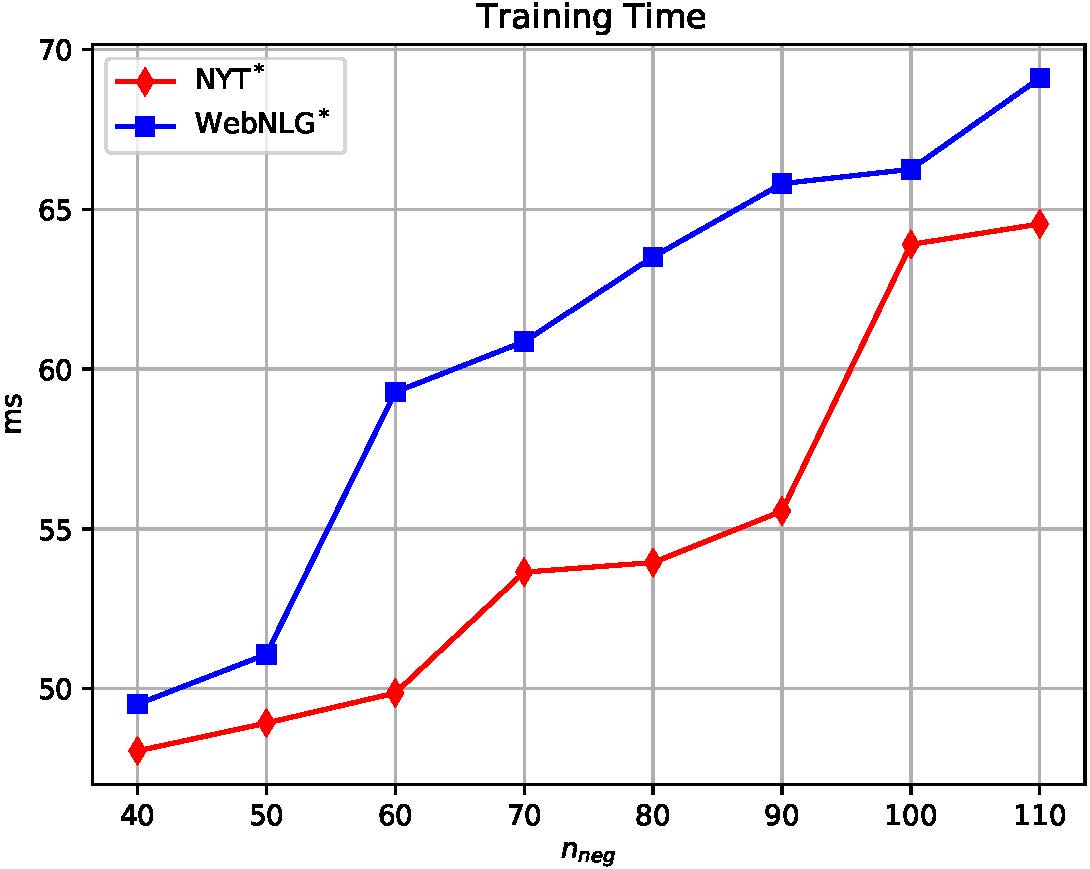}
		}
		\subfigure[GPU Memory]{
		\label{speed}
		\includegraphics[width = 5.2cm]{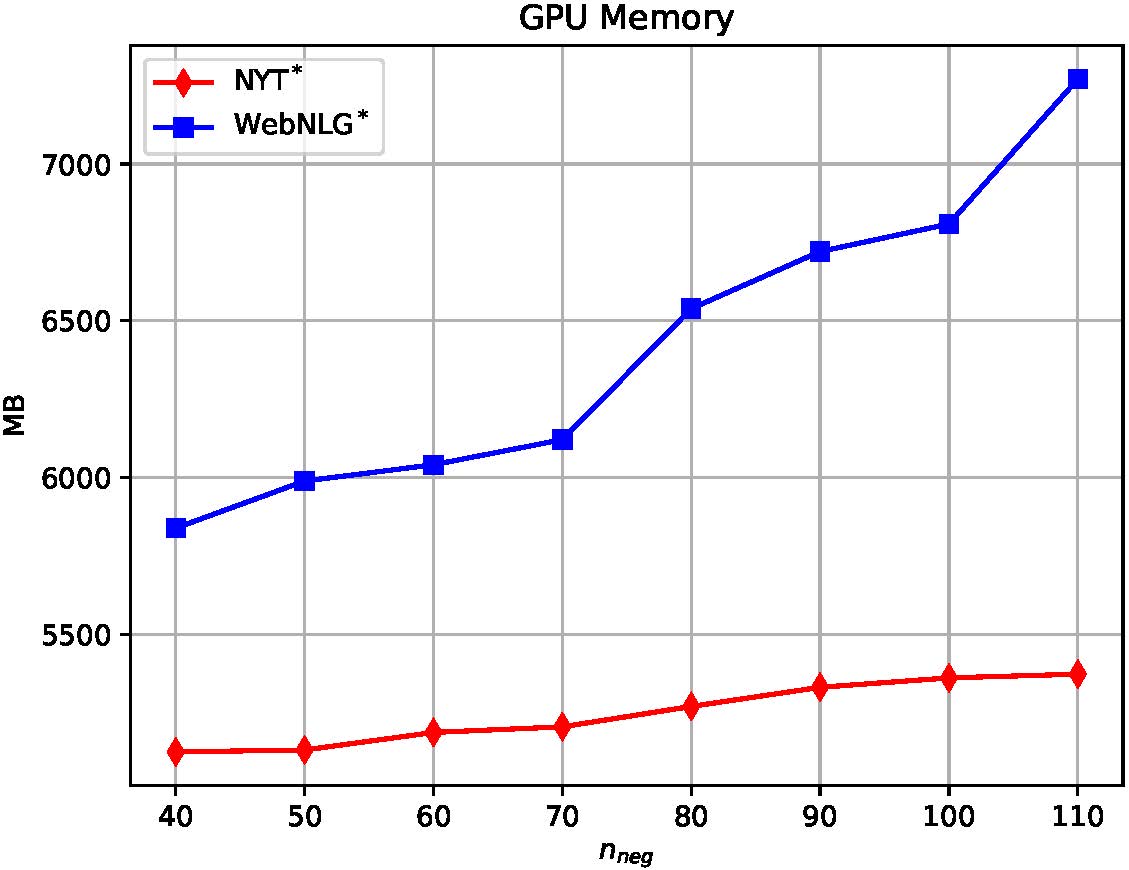}
		}
		\subfigure[F1-score]{
		\label{speed}
		\includegraphics[width = 5.2cm]{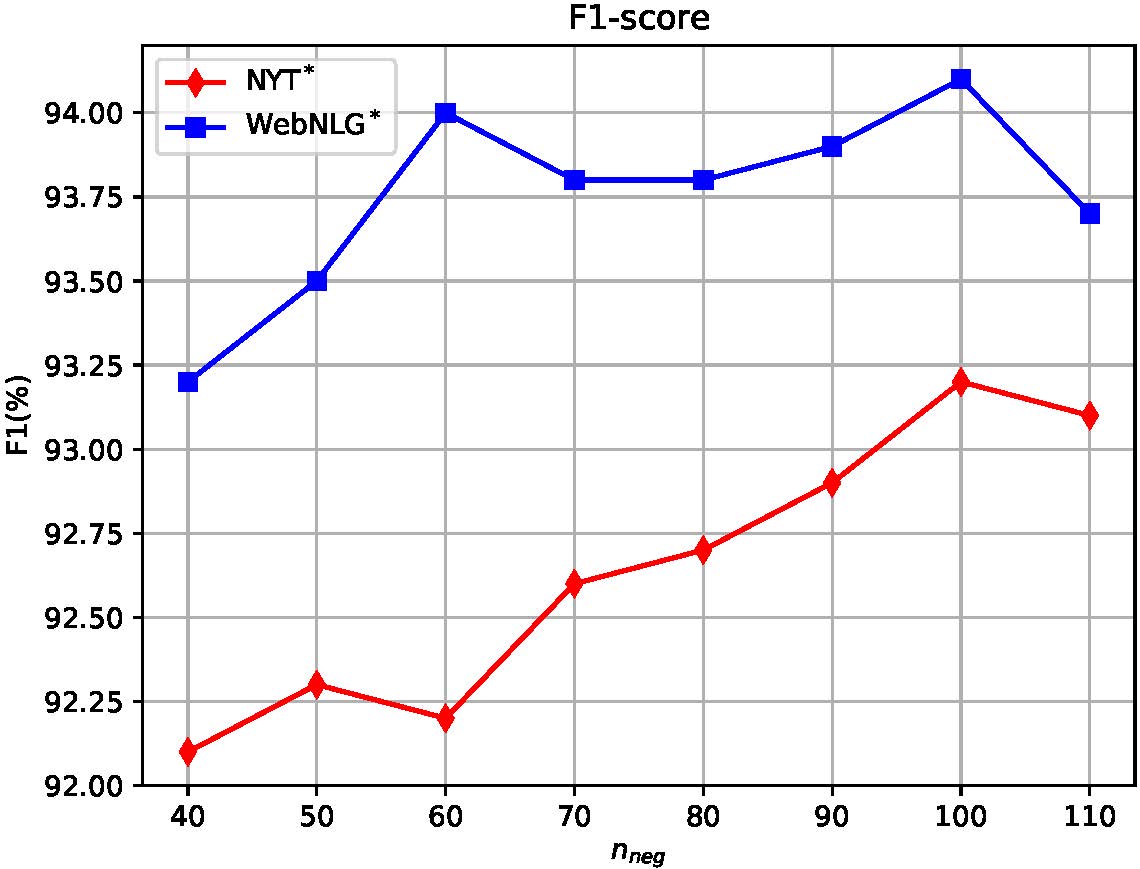}
		}
		\caption{The influence of different $n_{neg}$ in NYT$^*$ and WebNLG$^*$. Training time (ms) means the average time required to train one mini-batch, GPU memory (MB) is the average GPU memory required to train one epoch.}
		\label{fig:n_negative}
	\end{figure*}
	
	\begin{table}[t]
		\centering
		\setlength\tabcolsep{1.5mm}
		\renewcommand\arraystretch{1.1}
		\begin{tabular}{@{}cccccccc@{}}
			\toprule[2pt]
			\multirow{2}{*}{Model}  & \multirow{2}{*}{Element} & \multicolumn{3}{c}{NYT$^*$} & \multicolumn{3}{c}{WebNLG$^*$} \\ 
			\cmidrule(l){3-5} \cmidrule(l){6-8}
			&                          & Prec.   & Rec.  & F1    & Prec.    & Rec.   & F1     \\ \midrule
			\multirow{3}{*}{PRGC}  & $(h, t)$  & 94.0    & 92.3  & 93.1  & 96.0     & 93.4   & 94.7   \\
			& $r$   & 95.3    & 96.3  & 95.8  & 92.8     & 96.2   & 94.5   \\
			& $(h, r, t)$          & 93.3    & 91.9  & 92.6  & 94.0     & 92.1   & 93.0   \\ \midrule
			\multirow{3}{*}{DirectRel} & $(h, t)$  & \textbf{94.1}    & \textbf{93.2}  & \textbf{93.7}  & 95.8    & \textbf{95.9}   & \textbf{95.8}\\
			& $r$                        & \textbf{97.3}    & \textbf{96.4}  & \textbf{96.9}  & \textbf{96.8}     & \textbf{96.7}   & \textbf{96.7}   \\
			& $(h, r, t)$      & \textbf{93.7}    & \textbf{92.8}  & \textbf{93.2}  & \textbf{94.1}     & \textbf{94.1}   & \textbf{94.1}   \\ 
			\bottomrule[2pt]
		\end{tabular}
	\caption{Results on triple elements. $(h, t)$ denotes the entity pair and $r$ means the relation.}
	\label{tab:subtask}
	\end{table}
	
	\subsubsection{Results on Different Sub-tasks}
	Our DirectRel combines entity recognition and relation classification into a one-step bipartite graph link operation, which can better capture the interactions between the two sub-tasks. Furthermore, the one-step extraction logic protects the model from cascading errors and exposure bias.
	To verify such properties, we further explore the performance of DirectRel on the two sub-tasks. We select PRGC as baseline because (1) it is one of the state-of-the-art triple extraction models, and (2) it is powerful in relation judgement and head-tail alignment \cite{zheng-2021-prgc}. The results are shown in Table \ref{tab:subtask}.
	It can be found that DirectRel outperforms PRGC on all test instances except the precision of entity-pair recognition on WebNLG$^*$.
	This verifies our motivation again, that is, integrating entity recognition and relation extraction into a one-step extraction process can effectively enhance the correlation between the two tasks and improve their respective performance.		
	
	\subsubsection{Parameter Analysis}
	The most important hyper-parameter of our model is the number of negative samples $n_{neg}$, which aims to balance the convergence speed and generalization performance. In the following, we analyze the impact of $n_{neg}$ with respect to \textit{Training Time}, \textit{GPU Memory}, and \textit{F1-score} on  NYT$^*$ and WebNLG$^*$, the results are shown in Figure \ref{fig:n_negative}.
		
	It can be observed that with the increase of $n_{neg}$, the training time, GPU memory and F1-score on the two datasets show an upward trend, which is inline with our common sense. 
	Among them, the training time and GPU memory of our model on WebNLG$^*$ are significantly higher than that on NYT$^*$, the reason is that WebNLG$^*$ contains much more relations than NYT$^*$ (171 vs 24).
	Another interesting observation is that as $n_{neg}$ increases, the model performance shows a trend of increasing first and then decreasing.
	This phenomenon suggests that moderate and sufficient negative samples are beneficial for model training.
	
	\begin{table}[t]
	\centering
	\setlength\tabcolsep{10mm}
	\renewcommand\arraystretch{1.2}
        \begin{tabular}{@{}cc@{}}
		\toprule[2pt]
		Type & Distribution \\ 
		\midrule
		Span Splitting Error  & 35.5\%       \\
		Entity Not Found  & 19.4\%       \\
		Entity Role Error  & 45.1\%       \\ 
		\bottomrule[2pt]
		\end{tabular}
		\caption{Distribution of three entity recognition errors on WebNLG.}
		\label{tab:error}
	\end{table}
	
	\subsubsection{Error Analysis}
	Our DirectRel does not have an explicit process of entity boundary identification, so what is the main reason for the error of entity recognition in our method? To answer this question, we further analyze the types of entity errors on WebNLG and present the distribution of three errors:
	span splitting error, entity not found, entity role error in Table \ref{tab:error}. The proportion of ``span splitting error" is relatively small, which proves the effectiveness of directly extracting triples through link prediction on a directed ``head $\rightarrow$ tail" bipartite graph.
	Besides, the ``entity role error" is the most challenging to our method. The primary reason is that we ignore the contextual information of entities during triple extraction. We leave this issue for future work. 
	
\section{Conclusion}
	In this paper, we focus on addressing the error accumulation problem in existing relational triple extraction methods, and propose a one-step bipartite graph linking based model, named DirectRel, which is able to directly extract relational triples from unstructured text without specific processes of determining the start and end position of entities.
	Experimental results on two widely used datasets demonstrate that our model performs better than state-of-the-art baselines, especially for complex scenarios of different overlapping patterns and multiple triples.

\section{Acknowledgments}
The work is supported by National Key R\&D Plan (No. 2020AAA0106600), National Natural Science Foundation of China (No. U21B2009, 62172039, 61732005, 61602197 and L1924068 ), the funds of Beijing Advanced Innovation Center for Language Resources (No. TYZ19005).

\setlength{\baselineskip}{10.5pt}
\bibliographystyle{named}
\bibliography{ijcai22}

\clearpage

\section*{Appendix}

\subsection*{A \quad Overlapping Patterns}

\begin{figure}[h]
	\centering
	\includegraphics[width=1\columnwidth]{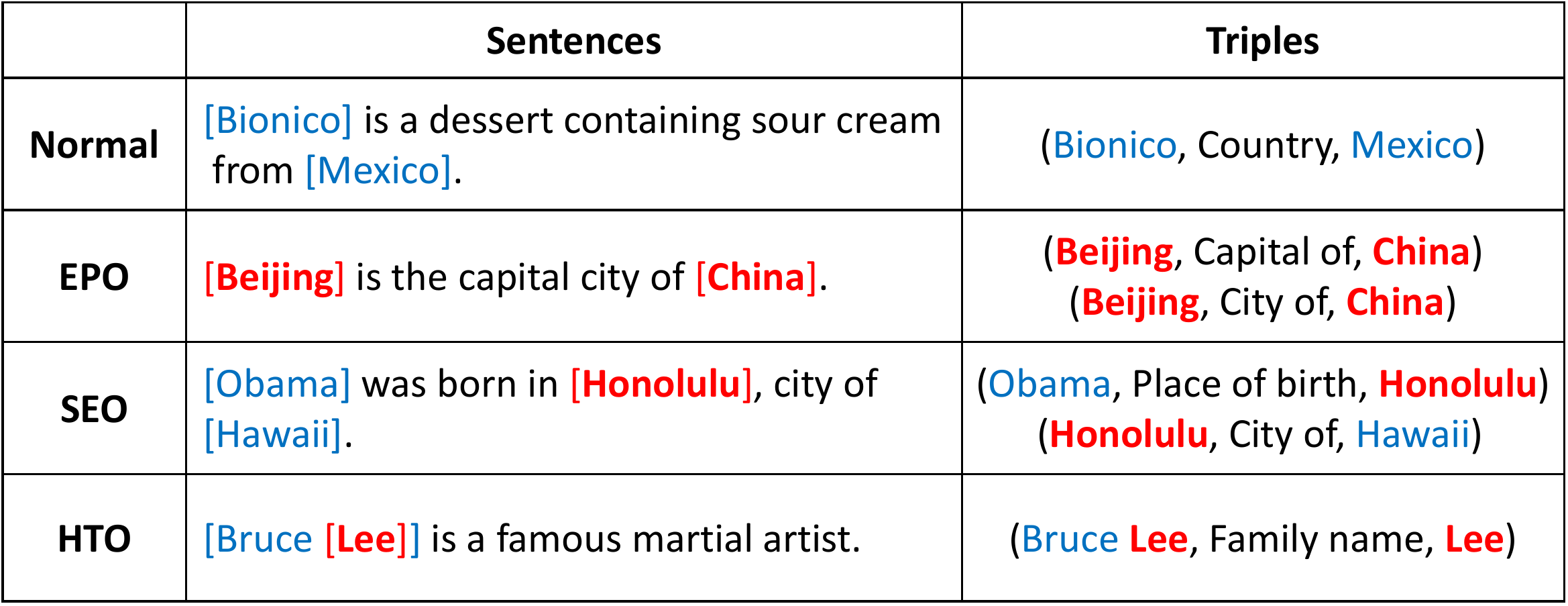} 
	\caption{
			Examples of \textit{Normal}, \textit{EPO}, \textit{SEO} and \textit{HTO} patterns. The overlapping entities are marked in red.
		}
	\label{fig:overlapping}
\end{figure}

As shown in Figure \ref{fig:overlapping}, sentences may contain \textit{EntityPairOverlap} (EPO), \textit{SingleEntityOverlap} (SEO) and \textit{HeadTailOverlap} (HTO) triples.
EPO means an entity pair holds multiple relations.
SEO denotes two triples share one overlapped entity.
HTO indicates that the head entity and tail entity of a triple are partially or completely overlapped.

\subsection*{B \quad Entity Length Distribution}

\begin{figure*}[!t]
		\centering
		\includegraphics[width=17.5cm]{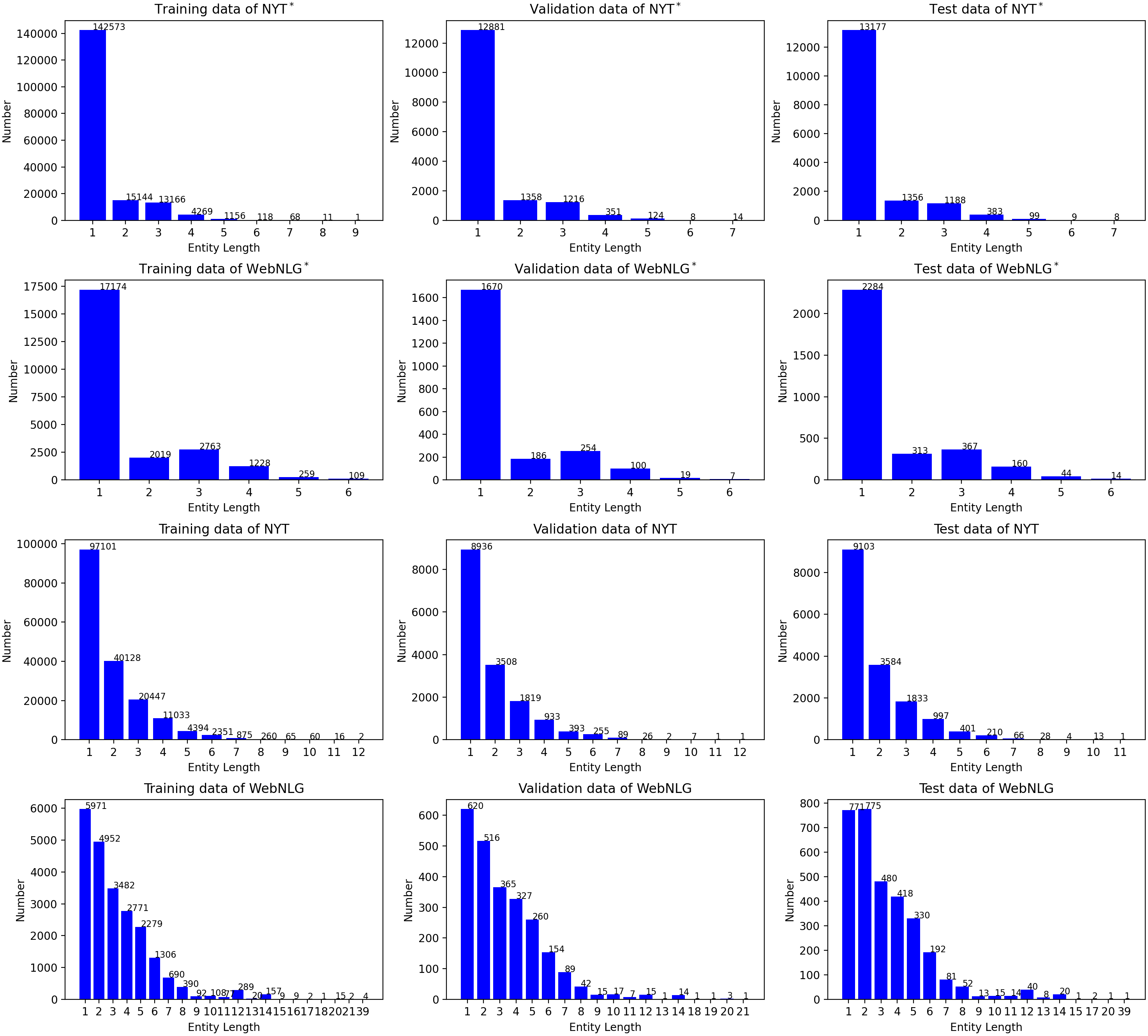} 
		\caption{
			   Entity length distribution of datasets using byte pair encoding.
			}
		\label{fig:distribution}
\end{figure*}

The entity length distribution of NYT$^*$, WebNLG$^*$, NYT and WebNLG are shown in Figure \ref{fig:distribution}.
When setting the max length of candidate entities $C$ , we mainly consider two factors:
(1) The generated candidate entities should contain as many correct entities as possible.
(2) The number of negative entities should not be too large.

Therefore, the max length of candidate entities $C$ is set to 9, 6, 12, 21 during training; 7, 6, 12, 21 during validation.
Strictly speaking, the distribution of entities in the test set is unknown at the time of model inference.
Therefore, during inference, we make adjustments based on the $C$ in the training set, and the final $C$ is set to 7, 6, 11, 20 during testing on NYT$^*$, WebNLG$^*$, NYT and WebNLG respectively.

\subsection*{C \quad Model Efficiency}

We randomly select $n_{neg}$ negative entities during training, while predict links for all candidate entities during inference. Therefore, we evaluate the model efficiency with respect to inference time of the token-pair linking method TPLinker in NYT$^*$, WebNLG$^*$, NYT, WebNLG, and the results are shown in Table \ref{tab:efficiency}.

\begin{table}[h]
\centering
	\setlength\tabcolsep{2mm}
	\renewcommand\arraystretch{1.1}
	\begin{tabular}{@{}ccccc@{}}
		\toprule[2pt]
		Model & NYT$^*$ &WebNLG$^*$ & NYT & WebNLG   \\ 
		\midrule
		TPLinker    & 46.2  & 40.1 & 45.1 & 42.2 \\
	    DirectRel     & 56.2  & 40.3 & 57.9 & 38.4 \\
		\bottomrule[2pt]
		\end{tabular}
	\caption{ The inference time (ms) of DirectRel and TPLinker to predict triples of one sentence.}
	\label{tab:efficiency}
\end{table}

From Table \ref{tab:efficiency} we can see that the inference speed of our DirectRel is competitive with TPLinker. The reason behind this are as follows: (1) Although our model needs to process more samples, its well-designed architecture is much simpler than TPLinker. (2) TPLinker requires a complex decoding algorithm to decode triples from output matrices, while the output of our model is directly the indices of entity spans and relations. 

Therefore, compared with the number of candidate entities, the model architecture and processing logic are more essential for the inference efficiency, and such number of candidate entities is not a bottleneck for the application of our model.

\end{document}